\def\year{2021}\relax
\documentclass[letterpaper]{article} 
\usepackage{aaai21}  
\usepackage{times}  
\usepackage{helvet} 
\usepackage{courier}  
\usepackage[hyphens]{url}  
\usepackage{graphicx} 
\usepackage{graphicx}  
\usepackage{natbib}  
\usepackage{caption} 
\usepackage{amsmath}
\usepackage{amssymb}

\title{AttaNet: Attention-Augmented Network for Fast and Accurate Scene Parsing }
\author{

    Qi Song,\textsuperscript{\rm 1,}\textsuperscript{\rm 3} 
    Kangfu Mei,\textsuperscript{\rm 1,}\textsuperscript{\rm 2}
    Rui Huang,\textsuperscript{\rm 1,}\textsuperscript{\rm 2}\thanks{Rui Huang is the corresponding author.}
    \\
}
\affiliations{

    \textsuperscript{\rm 1}Shenzhen Institute of Artificial Intelligence and Robotics for Society\\
    \textsuperscript{\rm 2}The Chinese University of Hong Kong, Shenzhen\\
    \textsuperscript{\rm 3}Jilin University\\

    \{songqi, ruihuang\}@cuhk.edu.cn,
    kangfumei@link.cuhk.edu.cn

}

\begin{document}

\maketitle

\begin{abstract}
    Two factors have proven to be very important to the performance of semantic segmentation models: global context and multi-level semantics. However, generating features that capture both factors always leads to high computational complexity, which is problematic in real-time scenarios. In this paper, we propose a new model, called Attention-Augmented Network (AttaNet), to capture both global context and multi-level semantics while keeping the efficiency high. AttaNet consists of two primary modules: Strip Attention Module (SAM) and Attention Fusion Module (AFM). Viewing that in challenging images with low segmentation accuracy, there are a significantly larger amount of vertical strip areas than horizontal ones, SAM utilizes a striping operation to reduce the complexity of encoding global context in the vertical direction drastically while keeping most of contextual information, compared to the non-local approaches. Moreover, AFM follows a cross-level aggregation strategy to limit the computation, and adopts an attention strategy to weight the importance of different levels of features at each pixel when fusing them, obtaining an efficient multi-level representation. We have conducted extensive experiments on two semantic segmentation benchmarks, and our network achieves different levels of speed/accuracy trade-offs on Cityscapes, e.g., 71 FPS/79.9\% mIoU, 130 FPS/78.5\% mIoU, and 180 FPS/70.1\% mIoU, and leading performance on ADE20K as well.
\end{abstract}

\section{Introduction}

\begin{figure}[t]
    \centering
    \includegraphics[width=0.95\columnwidth]{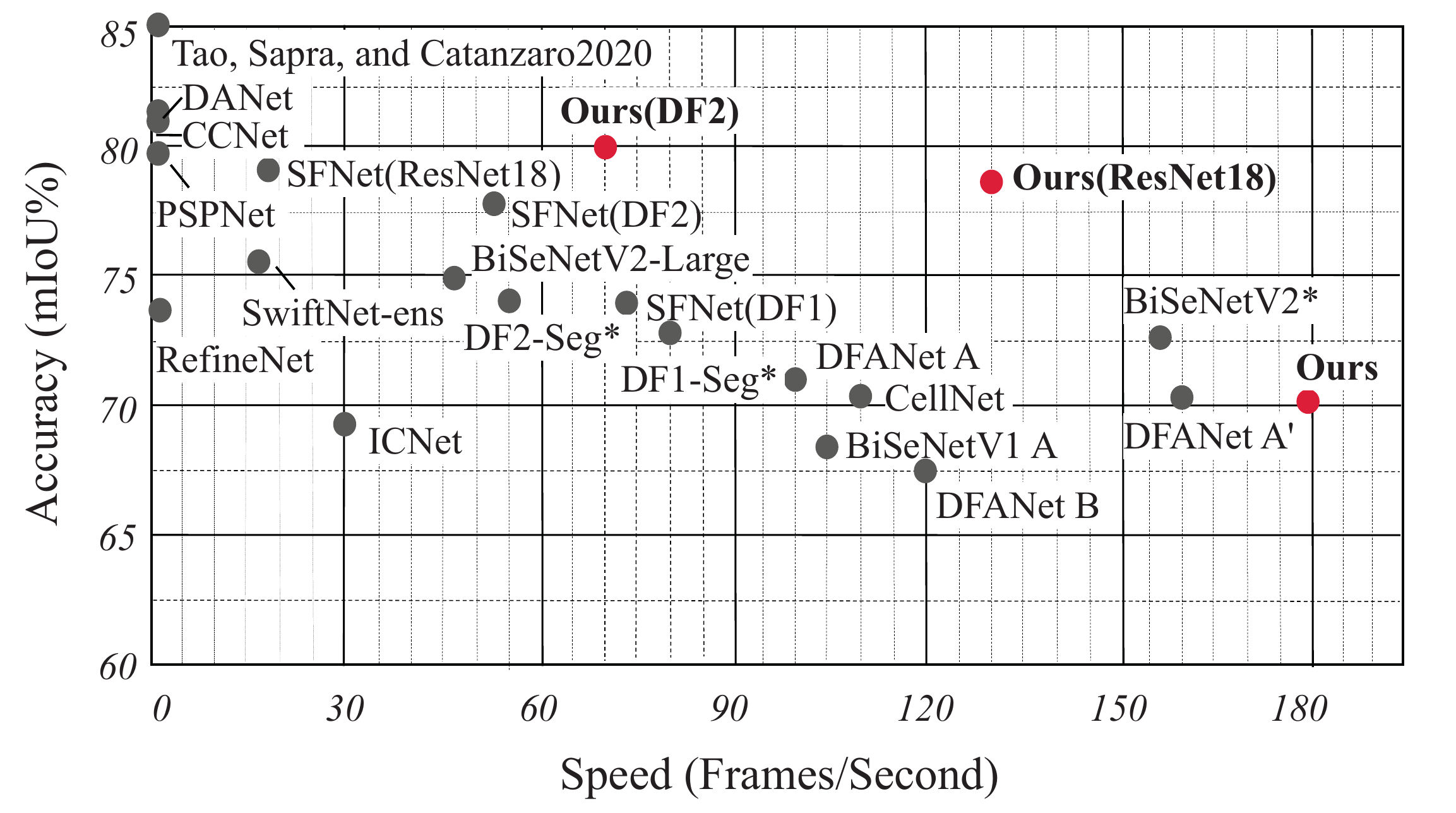}
    \caption{Inference speed and mIoU performance on the Cityscapes test set. Our method is marked as red points, while grey dots represent other methods. $\ast$ indicates that the model uses TensorRT for acceleration.}
    \label{fig1}
\end{figure}

Scene parsing, also known as semantic segmentation, predicts dense labels for all pixels in an image. As one of the fundamental tasks in computer vision, it has various applications in the fields of autonomous driving, video surveillance, robot sensing, and so on, many of which have a high demand for both segmentation accuracy and inference speed.

To achieve high accuracy, segmentation models need to generate features with global context information and multi-level semantics, both of which are known to be important factors in scene parsing. Global scene clues are typically captured via heavy networks with sizable receptive fields, e.g., PSPNet \cite{zhao2017pyramid}, DANet \cite{fu2019dual}, and AlignSeg \cite{huang2020alignseg} use ResNet101 \cite{he2016identity} as the backbone network. Besides, multi-level representations rely on both semantic information in high-level features and spatial details in low-level features. Nevertheless, both of them almost always require huge computation which is problematic to real-time applications.

On the other hand, in order to accelerate models for real-time scenarios, many state-of-the-art methods adopt light-weight backbone networks or restrict the size of input images. Though greatly boosting the inference speed, their accuracy is still unsatisfying due to the compromise on the aforementioned two factors. To achieve a better speed/accuracy trade-off, we propose Attention-Augmented Network (AttaNet) which can capture both global context and multi-level representations while keeping high computational efficiency.

In order to capture non-local contextual information with limited computation, we started by investigating one of the most commonly used approaches in scene parsing, the self-attention mechanism \cite{cheng2016long, vaswani2017attention}, which is capable of capturing long-range dependencies. Yet we can observe that these self-attention based models need to generate huge attention maps which are computationally expensive, e.g., the non-local block in Non-local Networks \cite{wang2018non} and the position attention module in DANet \cite{fu2019dual} both have a computational complexity of \(O((H\times W)\times (H\times W))\), where \(H\) and \(W\) donate the spatial dimensions of the input feature map. Therefore, many recent studies were proposed to achieve the same goal in a more efficient way, e.g., CCNet \cite{huang2019ccnet} reduces the computational complexity to \(O((H\times W)\times (H+W-1))\). However, the computational overhead is still too high to meet the real-time requirement.

In this work, we address this challenge by proposing an efficient self-attention based module called Strip Attention Module (SAM). SAM is inspired by the segmentation results in previous works, from which we find that various networks all achieved the lowest accuracies in classes such as \textit{fence}, \textit{pole}, and \textit{train}, which are contextually consistent and robust in a specific direction. The usage of large square pooling kernels would corrupt the structural details of these classes and incorporate contaminating information from irrelevant regions. This motivates us to introduce a striping operation into the traditional self-attention method, which can reduce the size of the attention map and also strengthen the directional consistency. Specifically, viewing that for classes with low accuracy there are a significantly larger amount of vertical strip areas than horizontal ones, SAM utilizes a striping operation to encode the global context in the vertical direction and then harvests long-range relations along the horizontal axis. By applying SAM, each position in the feature map is connected with pixels in different column spaces, and the computational complexity is reduced to \(O((H\times W)\times W)\). Besides, SAM can be trivially modified to perform horizontal striping for different purposes.

Moreover, we investigate how to generate multi-level representations for each pixel with negligible computational overhead. In mainstream semantic segmentation architectures, the feature fusion method is used to incorporate multi-level semantics into encoded features. Here we choose the cross-level aggregation architecture for its high efficiency. However, we find that multi-level features have different properties, e.g., high-level features encode stronger semantics while low-level features capture more spatial details. Simply combining those features would limit the effectiveness of information propagation. To mitigate this issue, we propose an Attention Fusion Module (AFM) which adopts an attention strategy that learns to weight multi-level features at each pixel location with minimal computation. Besides, we only apply AFM between the last two stages of the backbone network to further improve the efficiency.

We conducted extensive experiments on the two most competitive semantic segmentation datasets, i.e., Cityscapes \cite{cordts2016cityscapes} and ADE20K \cite{Zhou_2017_CVPR}. Our approach achieves top performance on both of them. To illustrate the performance comparisons, we show the accuracy and inference time of diﬀerent networks on the Cityscapes dataset in Figure \ref{fig1}.

To summarize, our main contributions are three-fold:
\begin{itemize}
    \item We introduce a Strip Attention Module which is able to capture long-range dependencies with only slightly increased computational cost.
    \item We propose a novel Attention Fusion Module to weight the importance of multi-level features during fusion, which attains a multi-level representation effectively and efficiently.
    \item Not only did our network achieve the leading performance on Cityscapes and ADE20K, the individual modules can also be combined with different backbone networks to achieve different levels of speed/accuracy trade-offs. Specifically, our approach obtains 79.9\%, 78.5\%, and 70.1\% mIoU scores on the Cityscapes test set while keeping a real-time speed of 71 FPS, 130 FPS, and 180 FPS respectively on GTX 1080Ti.
\end{itemize}

\section{Related Work}
\subsubsection{Self-attention Model.}
Self-attention models can capture long-range dependencies and have been widely used in many tasks. Mou et al. \cite{mou2019relation} introduced two network units to model spatial and channel relationships respectively. OCNet \cite{Yuan2018OCNetOC} and DANet \cite{fu2019dual} use the self-attention mechanism to capture long-range dependencies from all pixels. However, these methods need to generate huge attention maps, which adds much computational overhead. To reduce the complexity of the self-attention mechanism, CCNet \cite{huang2019ccnet} leverages two criss-cross attention modules to generate sparse connections \((H+W+1)\) for each position. ACFNet \cite{zhang2019acfnet} directly exploits class-level context to reduce the computation along channel dimensions. To capture long-range relations more effectively and efficiently, we introduce a striping operation in the Strip Attention Module. Different from the strip pooling used to enlarge the receptive field in work \cite{hou2020strip}, ours is designed to strengthen the contextual consistency in a specific direction while reduce the size of the affinity map.

\subsubsection{Multi-level Feature Fusion.}
Feature fusion is frequently employed in semantic segmentation to combine multi-level representations \cite{long2015fully, ronneberger2015u, lin2017refinenet, badrinarayanan2017segnet, fu2019adaptive}. For multi-level feature fusion, one solution is to adopt the multi-branch framework, e.g., ICNet \cite{zhao2018icnet} and BiSeNet series \cite{yu2018bisenet, yu2020bisenet} add an extra branch to remedy the lost spatial details in high-level features. To further boosts the inference speed, another type of methods \cite{li2019dfanet,li2020semantic} implement a cross-level feature aggregation architecture with less computation. Nevertheless, all these methods ignore the representation gap among multi-level features, which limits the effectiveness of information propagation. Recently, GFF \cite{Li2019GFFGF} uses gates to control information propagation, but ignores to limit the computation while maintaining effectiveness. In this regard, we propose the Attention Fusion Module which adopts a lightweight attention strategy to bridge the gap among multi-level features with high adaptability and efficiency.

\subsubsection{Real-time Segmentation.}
The goal of real-time semantic segmentation algorithms is to generate high-quality predictions while keeping high inference speed. ICNet proposes an image cascade network using multi-resolution images as input to raise efficiency. BiSeNetV2 introduces a detail branch and a semantic branch to reduce calculation. Both of them adopt shallow layers on the high-resolution image to speed up, while other branches have deep layers to obtain high-level semantics on low-resolution images. Besides, DFANet and LiteSeg \cite{emara2019liteseg} adopt a lightweight backbone to speed up the inference. Different from these approaches, our model can work with large backbone networks while reducing computational complexity and reserving both semantic and spatial information.

\section{Method}
\begin{figure}[t]
    \centering
    \includegraphics[width=0.95\columnwidth]{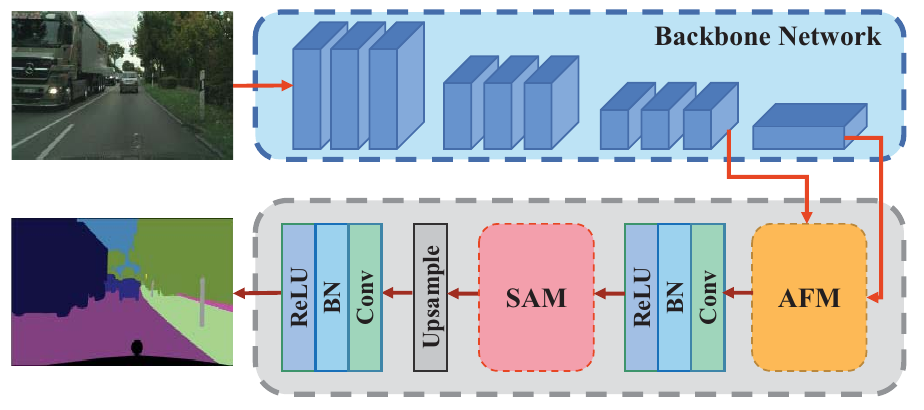} 
    \caption{Illustration of the overall architecture. In the figure, ResNet18  is used as the backbone for exemplar illustration.}
    \label{fig2}
\end{figure}

The overall network architecture of the proposed AttaNet is shown in Figure \ref{fig2}. As we can see, our AttaNet is a convolutional network that uses a cross-level aggregation architecture, which will be explained in the next subsection. Two key modules are then introduced respectively. To capture long-range relations efficiently, we propose the Strip Attention Module (SAM). And we introduce the Attention Fusion Module (AFM) where efficient feature aggregation is performed. Without loss of generality, we choose pre-trained ResNet \cite{he2016identity} from ImageNet \cite{russakovsky2015imagenet} as our backbone by removing the last fully-connected layer, and other CNNs can also be chosen as the backbone.

\subsection{Network Architecture}

In SAM, we add a Striping layer before the Affinity operation to get the strongest consistency along anisotropy or banded context. Then we utilize the Affinity operation to find out the long-range relations in the horizontal direction to further enhance the consistency. Furthermore, in AFM, we use an attention strategy to make the model focus on the most relevant features as needed, which bridges the representation gap between multi-level features and enables effective information propagation.

For explicit feature refinement, we use deep supervision to get better performance and make the network easier to optimize. We use the principal loss function to supervise the output of the whole network. Moreover, we add two specific auxiliary loss functions to supervise the output of the res3 block and AFM. Finally, we use a parameter \(\lambda\) to balance the principal loss and the auxiliary loss:
\[L=l_p+\lambda\sum_{i=1}^{K}l_i\,,\]
where \(l_p\)  is the principal loss of the final output. \(l_i\) is the auxiliary loss for the output of the res3 block and AFM. \(L\) is the joint loss. Particularly, all the loss functions are cross-entropy losses. \(K\) and \(\lambda\) are equal to 2 and 1 respectively in our implementation.

\begin{figure*}[t]
    \centering
    \includegraphics[width=0.9\textwidth]{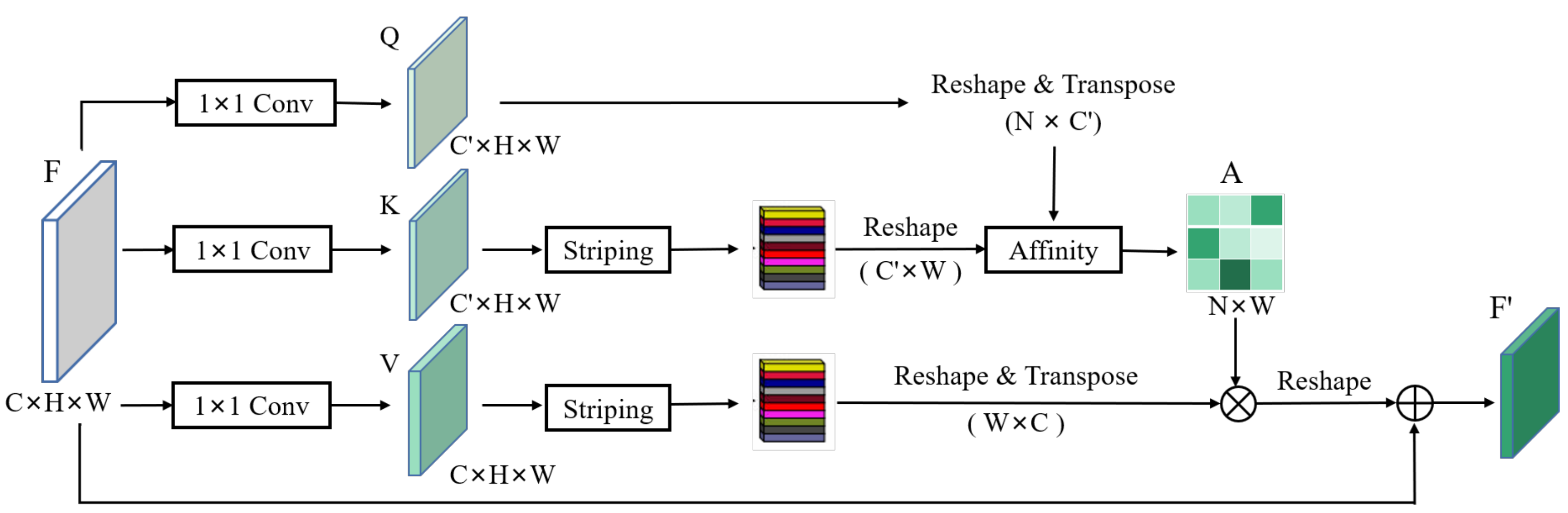}
    \caption{ The details of Strip Attention Module.}
    \label{fig4}
\end{figure*}

\subsection{Strip Attention Module}
In order to capture non-local contextual relations and also reduce the computational complexity in time and space, we introduce a module called Strip Attention Module. In particular, motivated by segmentation results in precious works \cite{fu2019dual,  zhang2019acfnet}, we apply a Striping operation to maintain contextual consistency along the vertical direction and further gather global affinity information between each pixel and banded structures along the horizontal axis. Figure \ref{fig4} gives the detailed settings of Strip Attention Module.

More precisely, given an input feature map \(F\in \mathbb{R}^{C\times H \times W}\), where \(C\) is the number of channels, \(H\) and \(W\) are the spatial dimensions of the input tensor. We first feed \(F\) into two convolution layers with \(1\times1\) filters to generate two new feature maps \(Q\) and \(K\) respectively, where \(\{Q, K\}\in \mathbb{R}^{C'\times H \times W}\). \(C'\) is less than \(C\) due to dimension reduction. We then apply a Striping operation on feature \(K\) to encode the global context representation in the vertical direction. Since the number of vertical strip areas is significantly larger than that of the horizontal ones in the natural images we are dealing with, the Striping operation represents average pooling with a pooling window of size \(H\times1\) in our work, and it can be extended to other directions for different purposes.

Then we reshape local features $Q$ and $K$ to $\mathbb{R}^{C'\times N}$ and $\mathbb{R}^{C'\times W}$ respectively, where $N=H\times W$ is the number of pixels.

After that, we perform an Affinity operation between \(Q^T\) and \(K\) to further calculate the attention map \(A\in \mathbb{R}^{N\times W}\) along the horizontal direction. The Affinity operation is defined as follows:
\[ A_{j, i}=\frac{exp(Q_{i}\cdot K_{j})}{\sum_{i=1}^N exp(Q_{i}\cdot K_{j})}\,,\]
where \(A_{j, i}\in A\) denotes the degree of correlation between \(Q_{i}\) and \(K_{j}\).

Meanwhile, we feed feature \(F\) into another convolutional layer with a kernel size of  \(1\times1\) to generate feature map \(V\in \mathbb{R}^{C\times H \times W}\). Similar to the above operation, for local feature \(V\) we can obtain a representation map in the vertical dimension whose spatial dimension is \(C\times 1\times W\) and reshape it to \(V\in \mathbb{R}^{C\times W}\). Then we perform a matrix multiplication between \(A\) and \(V^T\), and reshape the result to \(\mathbb{R}^{C\times H \times W}\). Finally, we perform an element-wise sum operation with the input feature map \(F\) to obtain the final output \(F'\in \mathbb{R}^{C\times H \times W}\) as follows:
\[F'_j=\sum_{i=1}^N A_{j, i}\cdot V_{i} + F_j\,,\]where \(F'_j\) is a feature vector in the output feature map \(F'\) at position \(j\). The contextual information is added to the input feature map \(F\) to augment the pixel-wise representation ability especially for banded structures.

The benefits of our SAM are three-fold. First, since the striped feature map is the combination of all pixels along the same spatial dimension, this gives strong supervision in capturing anisotropy or banded context. Second, we first ensure that the relationships between each pixel and all columns are considered, and then estimate the attention map along the horizontal axis, thus our network can generate dense contextual dependencies. Moreover, this module adds only a few parameters to the backbone network, and therefore takes up very little GPU memory.

\subsection{Attention Fusion Module}

\begin{figure}[t]
    \centering
    \includegraphics[width=0.95\columnwidth]{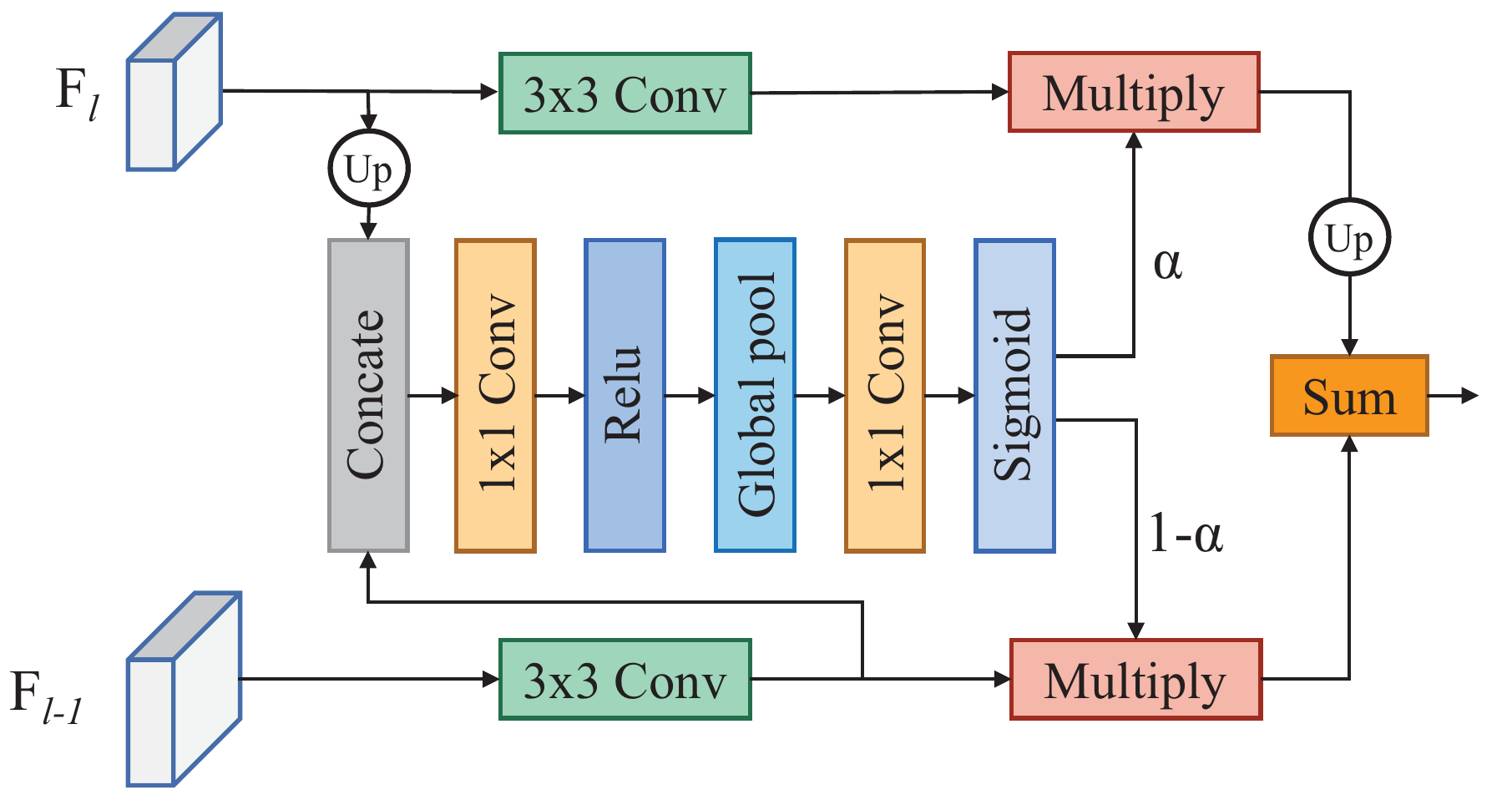} 
    \caption{ The details of Attention Fusion Module.}
    \label{fig3}
\end{figure}

As stated before, feature fusion is widely used for incorporating multi-level representations. The most commonly used approaches for aggregation are like works \cite{yu2018deep, chen2018encoder}, i.e., first upsampling \(F_l\) via bilinear interpolation and then concatenating or adding the upsampled \(F_l\) and \(F_{l-1}\) together. However, low-level features contain excessive spatial details while high-level features are rich in semantics. Simply aggregating multi-level information would weaken the effectiveness of information propagation. To address this issue, we introduce Attention Fusion Module which enables each pixel to choose individual contextual information from multi-level features in the aggregation phase.

The detailed structure of our Attention Fusion Module is illustrated in Figure \ref{fig3}.  Given two adjacent feature maps \(F_l\) and \(F_{l-1}\), we first upsample \(F_l\)  to the same size as \(F_{l-1}\) by the standard bilinear interpolation. Meanwhile, we feed \(F_{l-1}\) into a \(3\times3\) convolutional layer (with BN and ReLU). Then the upsampled  \(F_l\)  is concatenated with the local feature \(F_{l-1}\) , and the concatenated features are fed to a \(1\times1\) convolutional layer. After that, we leverage a global average pooling operation followed by a convolutional layer with kernel size of \(1\times1\) to predict the relative attention mask \(\alpha \). After obtaining two attention maps, we further perform pixel-wise product between masks and the predictions followed by pixel-wise summation among them to generate the final results, i.e.,
\[Output=Sum(Upsample(F_l)\cdot \alpha,  F_{l-1} \cdot (1-\alpha) )\,.\]
\quad This module employs the relative attention mask of adjacent features to guide the response of both features. This way, it bridges the semantic and resolution gap between multi-level features compared to the simple combination.

\section{Experiments}
To evaluate the proposed approach, we conducted extensive experiments on the Cityscapes dataset \cite{cordts2016cityscapes} and the ADE20K dataset \cite{Zhou_2017_CVPR}. Experimental results demonstrate that AttaNet obtains leading performance on both Cityscapes and ADE20K. In the following subsections, we first introduce the datasets and implementation details, and then we carry out a series of comparisons and ablation experiments on the Cityscapes dataset. Finally, we report our results on the ADE20K dataset.

\subsection{Datasets}
\subsubsection{Cityscapes.}
Cityscapes is a dataset for urban scene segmentation, which contains 5000 images with fine pixel-level annotations and 20000 images with coarse annotations. Each image has a resolution of \(1024\times2048\) and contains 19 classes of semantic labels. The 5000 images with fine annotations are further divided into 3 subsets of 2975, 500, and 1525 images for training, validation, and testing, respectively.

\subsubsection{ADE20K.}
ADE20K is a challenging scene parsing benchmark. The dataset contains 20K/2K images for training and validation which are densely labeled as 150 stuff/object categories.  Images in this dataset are from different scenes with more scale variations.

\subsection{Implementation Details}
Our model utilizes ImageNet pre-trained ResNet18 \cite{he2016identity} as the backbone. The last fully-connected layer is removed and the feature fusion method is applied between the output of the res4 block and res5 blocks.

\subsubsection{Training Settings.}
We train the network using standard SGD \cite{krizhevsky2012imagenet}. The mini-batch size is set to 16 and 32 for Cityscapes and ADE20K respectively. And we use the momentum of 0.9 and a weight decay of  \(5e^{(-4)}\). Similar to other works\cite{chen2017rethinking, yu2018learning}, we apply the `poly' learning rate policy in which the initial learning rate is set to \(1e^{(-2)}\) and decayed by \((1-\frac{iter}{max_iter})^{power}\) with power=0.9. The training images are augmented by employing random color jittering, random horizontal flipping, random cropping, and random scaling with 5 scales \{0.75, 1.0, 1.5, 1.75, 2.0\}. For Cityscapes, images are cropped into size of \(1024\times1024\), and the network is trained with 200k iterations. For ADE20K, crop size of \(512\times512\) and 250K training iterations are used for training.

\subsubsection{Inference.}
During the inference phase, we use a full image as input and follow the resizing method used in BiSeNetV2 \cite{yu2020bisenet}. For quantitative evaluation, the standard metric of mean pixel intersection-over-union (mIOU) is employed for accurate comparison and performance measuring, frames per second (FPS), number of float-point operations (FLOPs), and the number of model parameters are adopted for speed comparison.

\subsection{Experiments on Cityscapes}
\subsubsection{Comparisons with state of the art.}
\begin{table}[t]
    \centering
    \small
    \begin{tabular}{lcc}
        \hline
        Approach                                   & Backbone   & mIoU / FPS             \\
        \hline
        ICNet \cite{zhao2018icnet}                 & ResNet50   & 69.5 / 34              \\
        BiSeNetV1 A \cite{yu2018bisenet}           & Xception39 & 68.4 / 105.8           \\
        SwiftNet \cite{orsic2019defense}           & ResNet18   & 75.5 / 39.9            \\
        SwiftNet-ens \cite{orsic2019defense}       & ResNet18   & 76.5 / 18.4            \\
        DFANet A \cite{li2019dfanet}               & Xception A & 71.3 / 100             \\
        DFANet A' \cite{li2019dfanet}              & Xception B & 70.3 / \underline{160} \\
        DFANet B \cite{li2019dfanet}               & Xception B & 67.1 / 120             \\
        DF1-Seg\(^\dagger\) \cite{li2019partial}   & DF1        & 73.0 / 80.8            \\
        DF2-Seg\(^\dagger\) \cite{li2019partial}   & DF2        & 74.8 / 55              \\
        CellNet \cite{zhang2019customizable}       & -          & 70.5 / 108             \\
        BiSeNetV2\(^\dagger\) \cite{yu2020bisenet} & -          & 72.6 / 156             \\
        BiSeNetV2\(^\dagger\) \cite{yu2020bisenet} & -          & 75.8 / 47.3            \\
        SFNet \cite{li2020semantic}                & DF1        & 74.5 / 74              \\
        SFNet \cite{li2020semantic}                & DF2        & 77.8 / 53              \\
        SFNet \cite{li2020semantic}                & ResNet18   & 78.9 / 18              \\
        SFNet\(^\ddagger\) \cite{li2020semantic}   & ResNet18   & \textbf{80.4} / 18     \\
        FANet-18 \cite{Hu2020RealtimeSS}           & ResNet18   & 74.4 / 72              \\
        FANet-34 \cite{Hu2020RealtimeSS}           & ResNet18   & 75.5 / 58              \\
        \hline
        AttaNet                                    & -          & 70.1 / \textbf{180}    \\
        AttaNet (ResNet18)                         & ResNet18   & 78.5 / 130             \\
        AttaNet (DF2)                              & DF2        & \underline{79.9} / 71  \\
        \hline
    \end{tabular}
    \caption{Comparison on the Cityscapes test set with state-of-the-art real-time models. $\dagger$ indicates that the model is tested using TensorRT for acceleration. $\ddagger$ indicates the model uses Mapillary dataset for pretraining.}\smallskip
    \label{table5}
\end{table}

\begin{table}[t]
    \centering
    \begin{tabular}{ccccc}
        \hline
        SAM       & AFM       & mIoU (\%) & $\Delta$ a    & GFLOPs ($\Delta$) \\
        \hline
                  &           & 72.8      & -             & -                 \\
        \(\surd\) &           & 77.1      & 4.3$\uparrow$ & 0.185             \\
                  & \(\surd\) & 77.3      & 4.5$\uparrow$ & 0.336             \\
        \(\surd\) & \(\surd\) & 78.5      & 5.7$\uparrow$ & 0.521             \\
        \hline
    \end{tabular}
    \caption{Ablation study for the proposed modules on the Cityscapes validation set, where ResNet18 with feature aggregation architecture serves as the strong baseline.}\smallskip
    \label{table4}
\end{table}

In Table \ref{table5}, we provide the comparisons between our AttaNet and the state-of-the-art real-time models. Our method is tested on a single GTX 1080Ti GPU with a full image of \(1024\times2048\) as input which is resized into \(512\times1024\) in the model. Then we resize the prediction to the original size and the time of resizing is included in the inference time measurement, which means the practical input size is \(1024\times2048\). The speed is tested without any accelerating strategy and we only use train-fine data for training. As reported in Table \ref{table5}, we get 70.1\% mIoU with 180 FPS by stacking only eight convolution layers as the backbone network. As can be observed, the inference speed is significantly faster than that of the other models and the accuracy is comparable, which proves that even without heavy backbones our approach still achieves better performance than other approaches. Besides, our ResNet18 and DF2 based model achieves 130 FPS with 78.5\% mIoU and 71 FPS with 79.9\% mIoU respectively, which set the new state of the art on accuracy/speed trade-offs on the Cityscapes benchmark. It is worth mentioning that, with ResNet18 and DF2, the accuracy of our method even approaches the performance of the models that mainly focus on accuracy.

\begin{figure}[t]
    \centering
    \includegraphics[width=1\columnwidth]{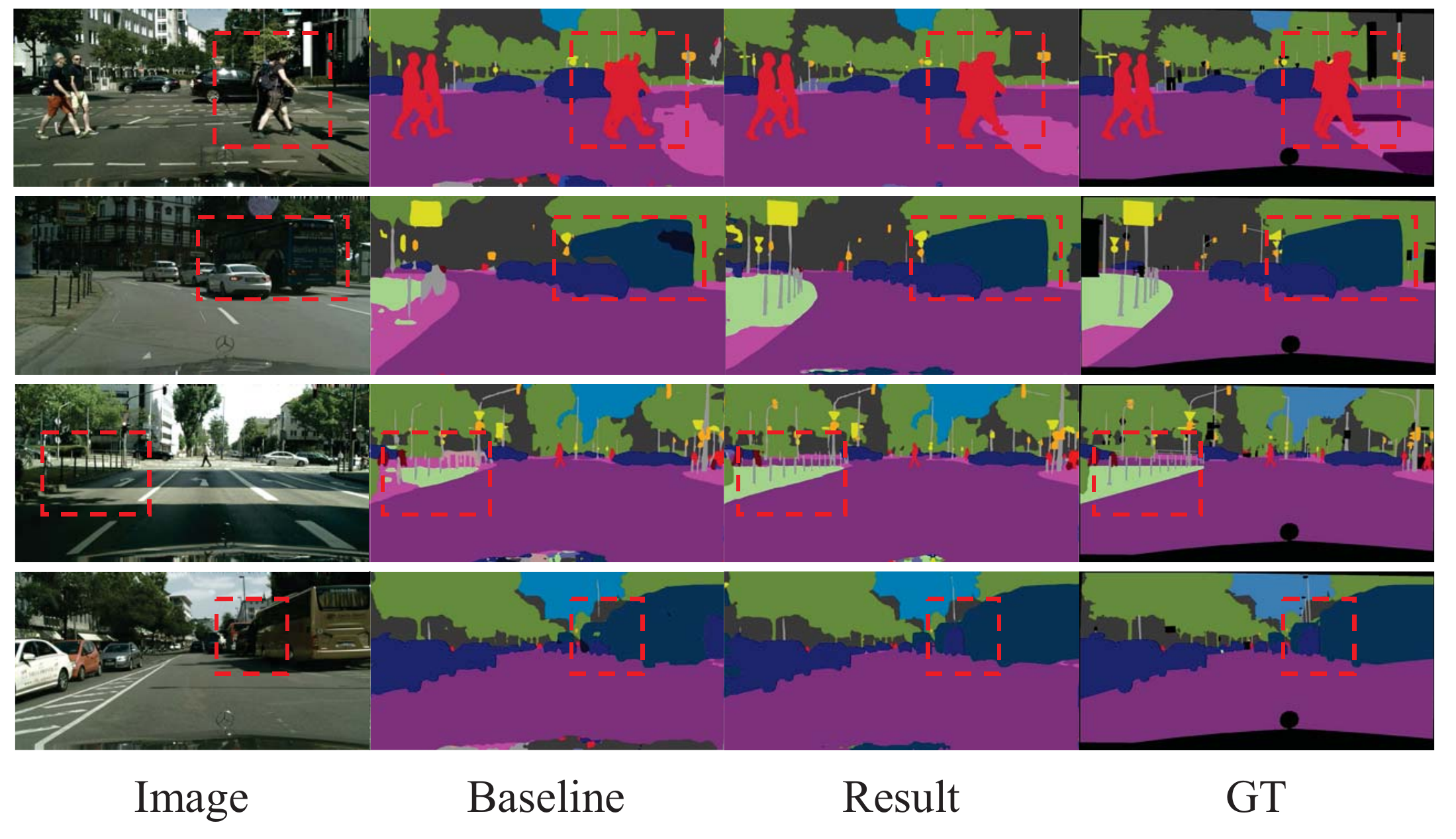}
    \caption{Visualization results of AttaNet on the Cityscapes validation set.}
    \label{fig5}
\end{figure}

To demonstrate the advantages of AttaNet, we provide the qualitative comparisons between AttaNet and the baseline in Figure \ref{fig5}. We use the red squares to mark the challenging regions. One can observe that the baseline network easily mislabels those regions but our proposed network is able to correct them, which clearly shows the effectiveness of AttaNet.

\begin{figure}[t]
    \centering
    \includegraphics[width=1\columnwidth]{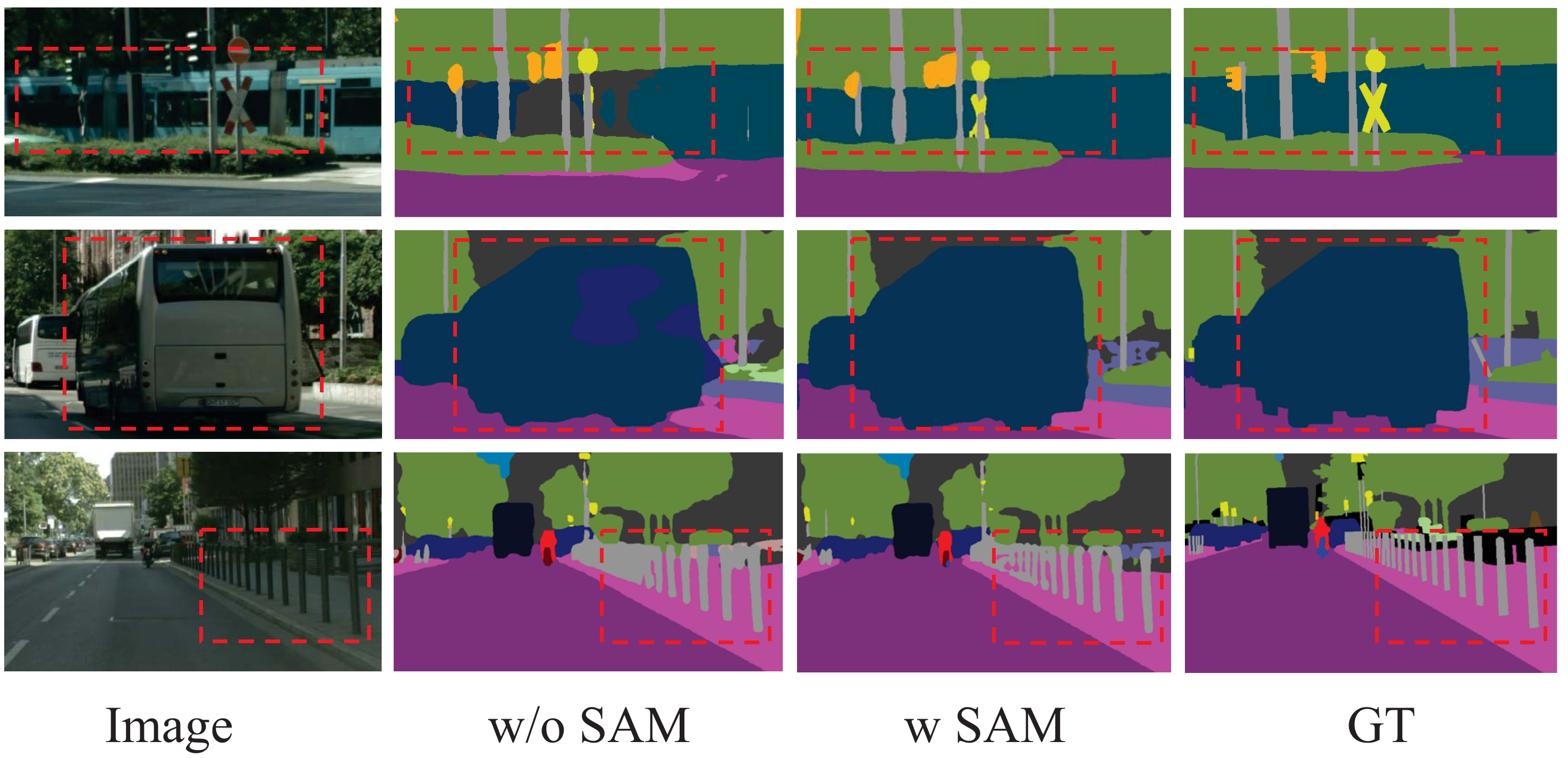} 
    \caption{Qualitative comparison between our approach w/o and w/ SAM on the Cityscapes validation set.}
    \label{fig7}
\end{figure}

\begin{figure}[t]
    \centering
    \includegraphics[width=1\columnwidth]{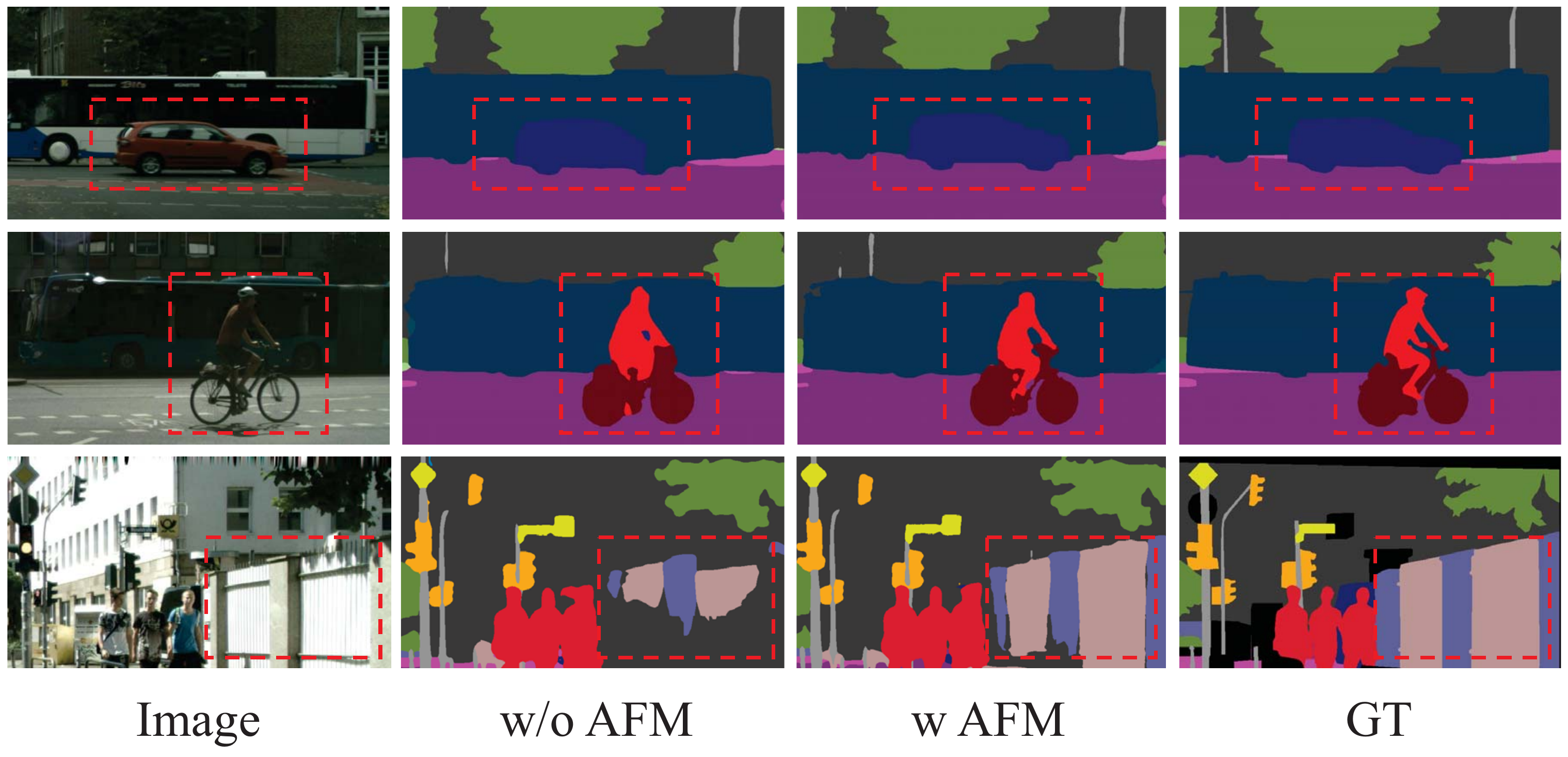} 
    \caption{ Qualitative comparison between our approach w/o and w/ AFM on the Cityscapes validation set.}
    \label{fig6}
\end{figure}
\begin{table}[t]
    \centering
    \small
    \begin{tabular}{lccc}
        \hline
        Approach                       & mIOU & GFLOPs & Memory \\
        \hline
        Baseline                       & 72.8 & -      & -      \\
        NL \cite{wang2018non}          & 78.1 & 3.357  & 334M   \\
        RCCA \cite{huang2019ccnet}     & 77.7 & 0.472  & 26M    \\
        EMA \cite{li2019expectation}   & 75.0 & 0.335  & 12M    \\
        \textbf{SAM-horizontal (Ours)} & 76.9 & 0.185  & 8M     \\
        \textbf{SAM-vertical (Ours)}   & 77.1 & 0.185  & 8M     \\
        \hline
        Addition (Baseline)            & 72.8 & -      & -      \\
        Concat                         & 73.7 & 0.336  & 10M    \\
        \textbf{AFM (Ours)}            & 77.3 & 0.336  & 12M    \\
        \hline
    \end{tabular}
    \caption{Comparison with other methods on the Cityscapes validation set, where ResNet18 with aggregation architecture is used as the baseline. GFLOPs (\(\Delta\)) and Memory usage (\(\Delta\)) are calculated for an input of \(1\times3\times1024\times1024\). }\smallskip
    \label{table2}
\end{table}

\begin{figure*}[t]
    \centering
    \includegraphics[width=0.98\textwidth]{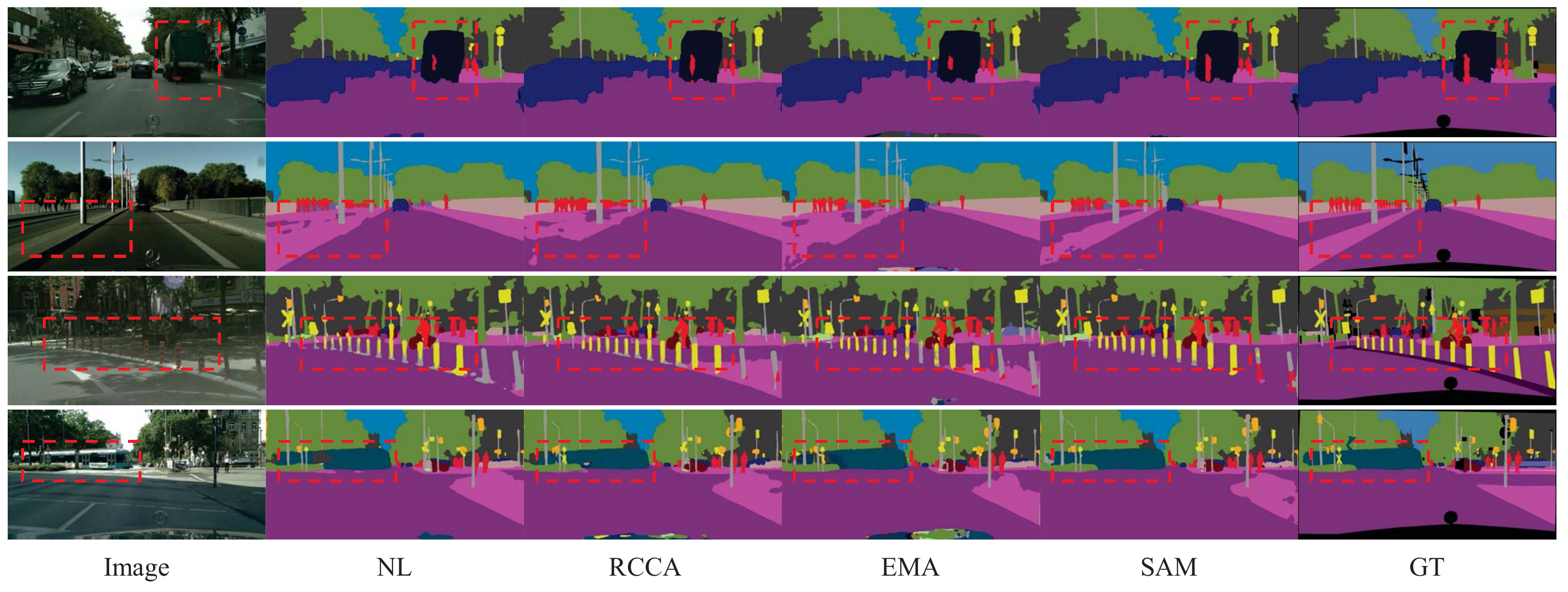}
    \caption{Qualitative comparison against different attention modules on the Cityscapes validation set.}
    \label{fig9}
\end{figure*}

\begin{figure}[t]
    \centering
    \includegraphics[width=1\columnwidth]{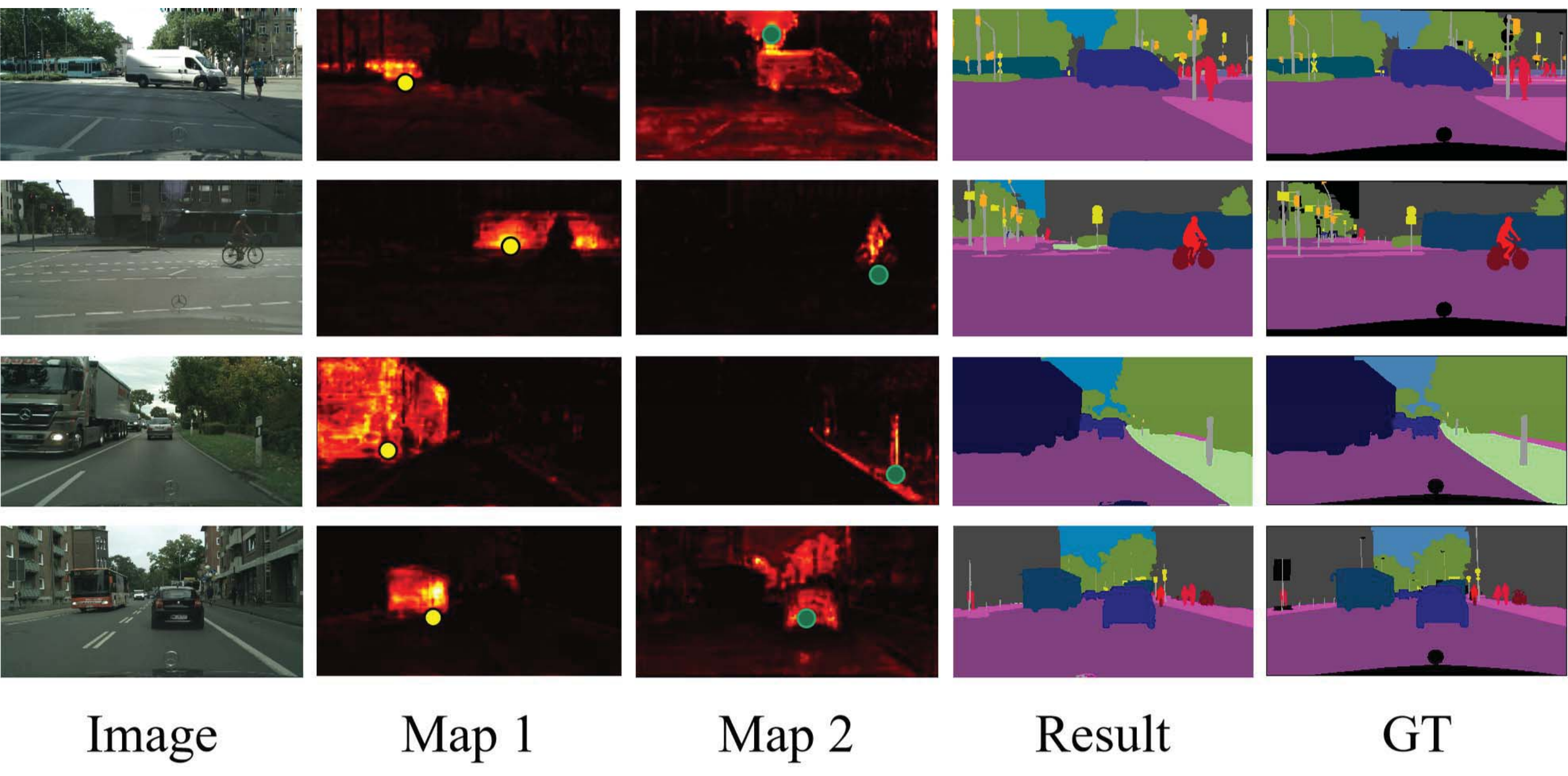} 
    \caption{Visualization results of SAM on Cityscapes val set.}
    \label{fig8}
\end{figure}

\subsubsection{Ablation study on proposed modules.}
To verify the effectiveness of the proposed modules, we first conduct ablation experiments on individual components, namely Strip Attention Module (SAM) and Attention Fusion Module (AFM). Without loss of generality, all results are obtained by training on the training set and evaluating on the validation set of Cityscapes. As shown in Table \ref{table4}, the baseline network achieves 72.8\% mIoU. By adding SAM, we get 77.1\% mIoU by an improvement of 4.3\%. Meanwhile, adding AFM brings 4.5\% mIoU gain. Finally, we append SAM and AFM together, which further improves mIoU to 78.5\%. All these improvements show that our modules bring great benefit to scene parsing. We also cropped some patches from some images in the Cityscapes val set, and show the comparison results in Figure \ref{fig7} and Figure \ref{fig6}. We superimposed red squares to mark those challenging regions. While other methods easily mislabel those areas, the proposed modules are able to rectify misclassification results. Also, we can observe that SAM generates more consistent segmentation inside large objects or along the banded areas, while AFM can exploit more discriminative context for each class, and that is probably why AFM achieves slightly higher performance than SAM does.

We further conducted a series of comparison experiments on other algorithmic aspects. Specifically, we adopt the amount of computation, Memory usage, and mIoU score for comparison. As shown in Table \ref{table2}, the top part compares the attention methods, and the bottom part compares the feature fusion methods. When given an input feature with a fixed size, SAM significantly reduces FLOPs by about 94.5\%, 60.8\%, and 44.8\% over NL, RCCA module in CCNet, and EMA unit in EMANet respectively. Compared with previous attention modules, our SAM achieves comparable segmentation performance while requiring significantly less GPU memory usage with both vertical striping (SAM-vertical) and horizontal striping (SAM-horizontal). Figure \ref{fig9} shows several qualitative comparisons, where SAM generates more consistent segmentation inside the banded objects. Moreover, we visualize the learned attention maps of SAM in Figure \ref{fig8}. For each input image, we select two columns (marked as yellow and green dots) and show their corresponding attention maps in columns 2 and 3 respectively. The last two columns are results from our AttaNet and the ground truth. We can find that SAM is able to capture long-range dependencies. Moreover, from the bottom part of Table \ref{table2}, we can observe that AFM achieves the best performance among the three methods with only slightly increased computation.

\subsubsection{Robustness on different backbones.}
To show the generalization ability of AttaNet, we further carry out a set of comparison experiments on adopting different backbone networks including both heavy and light-weight ones. Note that AttaNet can be easily inserted into various backbone networks. For light-weight backbones, we select ShuffleNetV2 \cite{ma2018shufflenet}, DF1, and DF2 \cite{li2019partial} as the representatives. For really deep networks, ResNet50 and ResNet101 \cite{he2016identity} are experimented on. Note that only ShuffleNetV2 and ResNet are pretrained on ImageNet. The comparison results are reported in Table \ref{table3}, which proves that our model can achieve considerably better mIoU on either heavy or light-weight backbones with only slightly increased computational cost.

\begin{table}[t]
    \centering
    \small
    \begin{tabular}{lcccc}
        \hline
        Backbone                             & mIoU & Params & GFLOPs \\
        \hline
        ResNet50 \cite{he2016identity}       & 73.4 & 41.51M & 171.36 \\
        +AttaNet                             & 81.2 & 53.62M & 176.91 \\
        ResNet101 \cite{he2016identity}      & 74.5 & 65.80M & 324.36 \\
        +AttaNet                             & 81.0 & 78.96M & 329.91 \\
        \hline
        ShuffleNetV2 \cite{ma2018shufflenet} & 67.7 & 2.55M  & 12.10  \\
        +AttaNet                             & 76.0 & 3.31M  & 12.59  \\
        DF1 \cite{li2019partial}             & 70.3 & 9.34M  & 25.92  \\
        +AttaNet                             & 78.0 & 10.43M & 26.44  \\
        DF2 \cite{li2019partial}             & 72.5 & 18.96M & 48.74  \\
        +AttaNet                             & 80.0 & 20.08M & 49.23  \\
        \hline
    \end{tabular}
    \caption{Ablation study on different backbones, where cross-level aggregation architecture is used as the baseline.}\smallskip
    \label{table3}
\end{table}

\begin{table}[h!]
    \centering
    \small
    \begin{tabular}{lcc}
        \hline
        Approach                          & Backbone  & mIoU/ GFLOPs           \\
        \hline
        PSPNet \cite{zhao2017pyramid}     & ResNet50  & 42.78 / 335.0          \\
        SFNet \cite{li2020semantic}       & ResNet50  & 42.81 / 151.1          \\
        \textbf{AttaNet}                  & ResNet50  & 41.79 / \textbf{116.3} \\
        \hline
        UperNet \cite{xiao2018unified}    & ResNet101 & 42.66 / -              \\
        PSPNet \cite{zhao2017pyramid}     & ResNet101 & 43.29 / 476.3          \\
        PSANet \cite{zhao2018psanet}      & ResNet101 & 43.77 / 529.3          \\
        SAC \cite{zhang2017sac}           & ResNet101 & 44.30 / -              \\
        EncNet \cite{zhang2018context}    & ResNet101 & 44.65 / -              \\
        SFNet \cite{li2020semantic}       & ResNet101 & 44.67 / 187.5          \\
        CFNet \cite{zhang2019co}          & ResNet101 & 44.82 / -              \\
        CCNet \cite{huang2019ccnet}       & ResNet101 & 45.22 / -              \\
        ACNet \cite{fu2019adaptive}       & ResNet101 & 45.90 / -              \\
        AlignSeg \cite{huang2020alignseg} & ResNet101 & 45.95 / -              \\
        \textbf{AttaNet}                  & ResNet101 & 43.71 / \textbf{150.5} \\
        \hline
    \end{tabular}
    \caption{Comparison on the ADE20K validation set with the state-of-the-art models.}\smallskip
    \label{table6}
\end{table}

\subsection{Experiments on ADE20K}
Table \ref{table6} reports the performance comparisons between AttaNet and the state-of-the-art models on the ADE20K validation set. Our approach achieves 41.79\% mIoU and 43.71\% mIoU respectively with much less computation.

\section{Conclusions }
In this paper, we focus on achieving a better speed/accuracy trade-off on the semantic segmentation task, and present an Attention-Augmented Network (AttaNet) for real-time scene parsing. First, we introduce Strip Attention Module to exploit long-range dependencies among all pixels. Particularly, by using the Striping operation, our network dramatically reduces the computation cost of the self-attention mechanism. Moreover, to attain a high-level and high-resolution feature map efficiently, we propose Attention Fusion Module which enables each pixel to choose private contextual information from multi-level features by utilizing the attention strategy. Experimental results show that AttaNet achieves outstanding speed/accuracy trade-offs on Cityscapes and ADE20K.

\section{Acknowledgments}
This work is supported in part by funding from Shenzhen Institute of Artificial Intelligence and Robotics for Society, and Shenzhen NSF JCYJ20190813170601651.

\bibliographystyle{aaai21}
\bibliography{ref}

\end{document}